\documentclass[10pt,twocolumn,letterpaper]{article}

\usepackage{iccv}
\usepackage{times}
\usepackage{epsfig}
\usepackage{graphicx}
\usepackage{amsmath}
\usepackage{amssymb}

\usepackage{siunitx}
\usepackage{cite}
\usepackage{ctable}
\usepackage{hhline}
\usepackage{booktabs}
\usepackage{subcaption}
\usepackage{csquotes}
\usepackage[accsupp]{axessibility}  

\usepackage[pagebackref=true,breaklinks=true,letterpaper=true,colorlinks,bookmarks=false]{hyperref}

\iccvfinalcopy 

\def\iccvPaperID{2350} 

\usepackage{float}
\usepackage{multirow}

\usepackage{url}

\usepackage{capt-of,etoolbox}

\makeatletter
\apptocmd\@maketitle{{\myfigure{}\par}}{}{}
\makeatother

\begin{document}

\title{Image Inpainting Models are Effective Tools \\
for Instruction-guided Image Editing\\
{\normalsize Solution for GenAI Media Generation Challenge Workshop @ CVPR}
}

\author{
Xuan Ju$^{1,2}$, 
Junhao Zhuang$^{1}$,
Zhaoyang Zhang$^{1*}$, 
Yuxuan Bian$^{1,2}$, 
Qiang Xu$^{2}$,
Ying Shan$^{1}$
\\
$^{1}$ARC Lab, Tencent PCG, $^{2}$The Chinese University of Hong Kong $^{*}$Project Lead\\
\small{\url{https://github.com/TencentARC/BrushNet/tree/main/InstructionGuidedEditing}}
}

\newcommand\myfigure{%
\centering
\vspace{-0.8cm}
   \includegraphics[width=1.\linewidth,trim={4pt 4pt 4pt 4pt}]{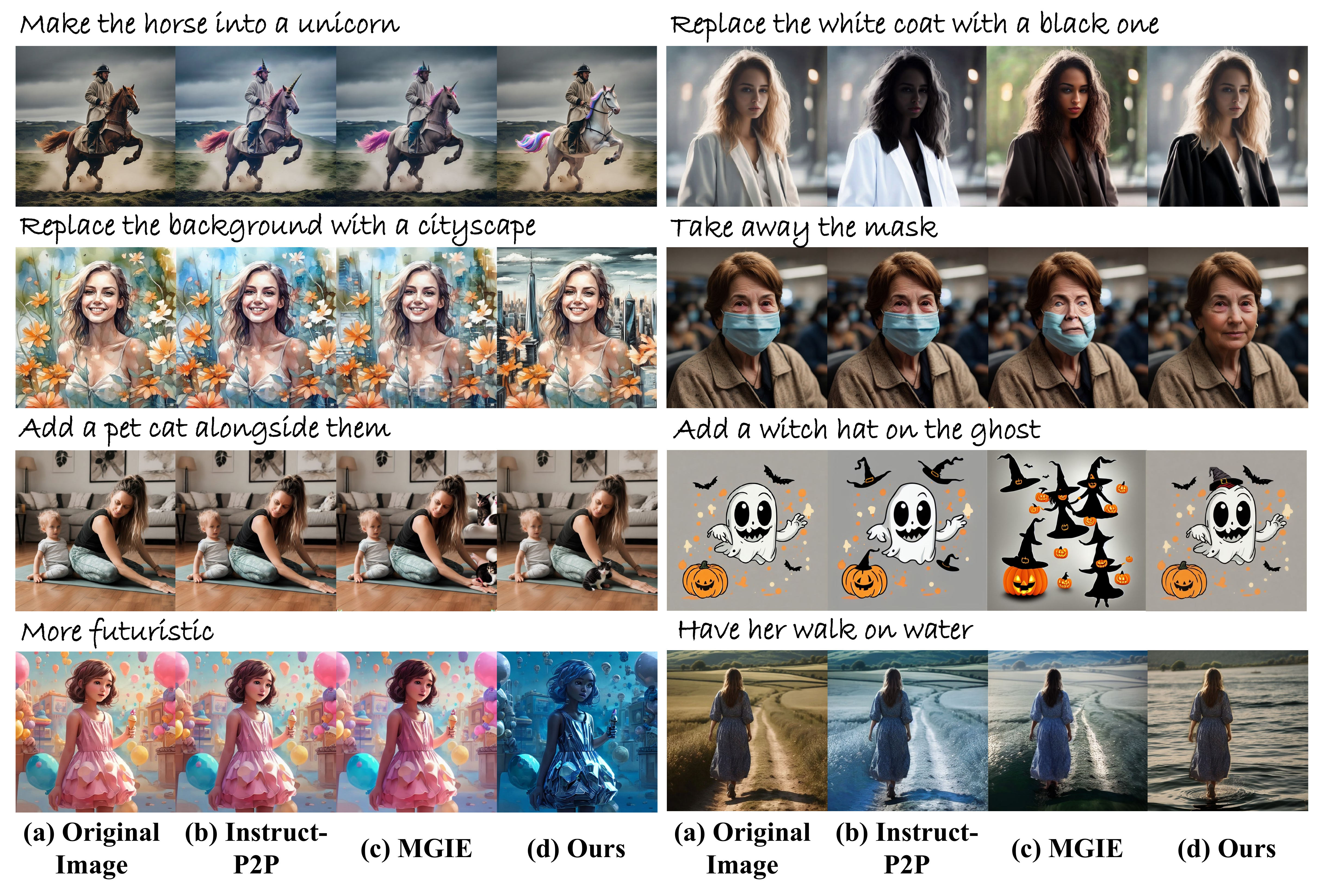}
\vspace{-0.9cm}
\captionof{figure}{\textbf{The comparison of Previous Text-Guided Image Editing Methods and Ours.} Text prompt is shown at the top of each group of images. We include editing category of local edit, background edit, global edit, addition, and remove.} 
\vspace{0.2cm}
\label{fig:teaser}
}
\maketitle

\thispagestyle{empty}

\begin{abstract}
\vspace{-0.5cm}
This is the technique report for the winning solution of the CVPR2024 GenAI Media Generation Challenge Workshop's Instruction-guided Image Editing track.
Instruction-guided image editing has been largely studied in recent years.
The most advanced methods, such as SmartEdit and MGIE, usually combine large language models with diffusion models through joint training, where the former provides text understanding ability, and the latter provides image generation ability.
However, in our experiments, we find that simply connecting large language models and image generation models through intermediary guidance such as masks instead of joint fine-tuning leads to a better editing performance and success rate. 
We use a 4-step process \textbf{IIIE} (\textbf{I}npainting-based \textbf{I}nstruction-guided \textbf{I}mage \textbf{E}diting): editing category classification, main editing object identification, editing mask acquisition, and image inpainting.
Results show that through proper combinations of language models and image inpainting models, our pipeline can reach a high success rate with satisfying visual quality.
\end{abstract}

\vspace{-2.5cm}
\section{Introduction}

With the rapid development of diffusion models, the field of text-guided image generation~\cite{ho2020denoising,song2020denoising,ju2023humansd,liu2023hyperhuman} has seen unprecedented progress in creating images with superior quality~\cite{Rombach_2022_CVPR}, diversity~\cite{emu}, and adherence to text guidance~\cite{li2024textcraftor}. However, in image editing tasks, which provide a source image and an editing instruction as input and expect a target image as output, we do not observe such success. This implies that the language understanding ability and image generation ability are not fully explored in editing tasks.

In the aim of applying image generation capabilities to image editing, previous methods have attempted two strategies: (1) collecting paired ``source image-instruction-target image'' data and fine-tuning diffusion models for editing tasks (\textit{e.g.}, InstructPix2Pix~\cite{instructpix2pix} and InstructDiffusion~\cite{instructdiffusion}), and (2) jointly fine-tuning Large Language Models (LLMs) and diffusion models to endow the diffusion models with a stronger understanding of text (\textit{e.g.}, SmartEdit~\cite{smartedit} and MGIE~\cite{mgie}). For the former strategy, due to the difficulty of collecting paired manually edited data, the training data are usually generated by LLMs and inference-based image editing methods (\textit{e.g.}, Prompt-to-Prompt~\cite{prompt2prompt} and Masactrl~\cite{masactrl}). Due to the low success rate and unstable editing quality of these inference-based image editing methods~\cite{directinversion}, the collected dataset is usually noisy and unreliable, which lead to an unsatisfying performance of the trained image editing model. For the latter strategy, a joint training of the LLMs and diffusion model usually make it hard to fully use the text understanding capability of LLMs. Although SmartEdit and MGIE have achieved better performance than previous solutions, we find that they still do not fully optimize the potential of LLMs and diffusion.

This technique report provides a different solution, connecting large language models and image generation models simply through intermediary guidance (\textit{e.g.}, edit objects and masks), and is the winning solution for The GenAI Media Generation Challenge (MAGIC). By detracting two models apart in an agent architecture, we find it easier to fully leverage the capabilities of both. Specifically, we use a 4-step process \textbf{IIIE} (\textbf{I}npainting-based \textbf{I}nstruction-guided \textbf{I}mage \textbf{E}diting): (1) editing category classification, (2) main editing object identification, (3) editing mask acquisition, and (4) image inpainting. Step (1)-(3) use LLMs and detection model to determine the editing type, edit object, edit masks, and target prompt. Then, step (4) perform image editing in the way of image inpainting, which fully use generation ability. In this way, step (1)-(3) use LLMs to extract information in instruction and summarize them to intermediary guidance that can be used by diffusion models.

Visual results show that the proposed IIIE substantially surpass previous instruction-guided image editing methods and other solutions in MAGIC considering visual quality and instruction faithfulness. 

\section{MAGIC}

The MAGIC hosts two challenge tracks: (1) text-to-image generation and (2) text-guided image editing. This technique report presents the solution for the second track, text-guided image editing. We list the instructions here:

\begin{center}
\fcolorbox{black}{gray!10}{\parbox{.98\linewidth}{
\textbf{Guidelines}

For text-guided image editing, we test the capacity of the model to change a given image's contents based on some text instructions. The specific type of instructions that we test for are the following:

\begin{itemize}
    \item Addition: Adding new objects within the images.
    \item Remove: Removing objects
    \item Local: Replace local parts of an object and later the object's attributes, i.e., make it smile
    \item Global: Edit the entire image, i.e., let's see it in winter
    \item Inpaint: Alter an object's visual appearance without affecting its structure
    \item Background: Change the scene's background
\end{itemize}
}}
\end{center}

\begin{center}
\fcolorbox{black}{gray!10}{\parbox{.98\linewidth}{

\textbf{Evaluation Protocol}

We leverage both human and automated evaluations. In human-based evaluations, we use human annotators. We mainly evaluate the following aspects:

\begin{itemize}
    \item Edit Faithfulness - whether the edited image follows the editing instruction
    \item Content Preservation - whether the edited image preserves the regions of the original image that should not be changed
     \item Overall Instruction Following - considering both edit faithfulness and content preservation, whether the edited image is artifact-free, keeping the core visual features of the original image, etc
\end{itemize}

On automatic evaluation, similar to the text to image track, we will leverage existing methods that have developed automatic metrics to help in assessing the outputs of the image based on the prompt and instruction.

To determine winners, we use automatic evaluation to help prune the total number of entries to 10 finalists. At 10, we would use human annotation and evaluation to determine the final winners. 

}}
\end{center}

More information about the detailed workshop information can be found at the official website page: \url{https://gamgc.github.io/}.

\begin{figure*}[htbp]
\centering
\includegraphics[width=1.0\linewidth]{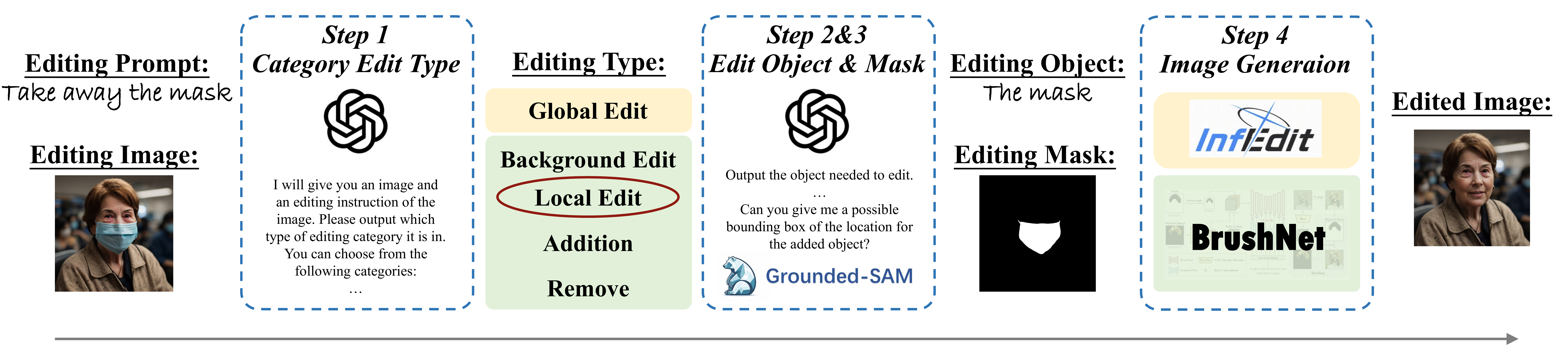}
\caption{\textbf{The pipeline of our proposed 4-step image editing process ITIE.}} 
\label{fig:main}
\end{figure*}

\section{Method}

Previous instruction-guided image editing methods include two categories: (1) diffusion model finetuning~\cite{instructpix2pix,instructdiffusion}, and (2)  jointly LLMs and diffusion model~\cite{smartedit,mgie}. 
These methods either not include LLMs in the model, or use joint fine-tuning of LLMs and diffusion models. Both of these strategies can not fully unleash the capabilities of LLMs, leading to a weak understanding of instructions.
Consequently, these methods show low success rates and unsatisfying results.
Contrary to these methods, we find that a simple tool-based combination of LLMs and diffusion models can lead to a much better visual results, coming from a full utilization of the language understanding ability of LLMs.

In this competition, we propose a 4-step process for instruction-based image editing, IIIE (Inpainting-based Instruction-guided Image Editing), as shown in Fig.~\ref{fig:main}.
Firstly, we use GPT4-o to categorize the current editing instructions into one of the editing categories: Local Edit, Background Edit, Global Edit, Addition, and Remove. 
Local Edit includes local changes such as replacing object (\textit{e.g.}, change a cat to a dog) and change the attribute of an object (\textit{e.g.}, change color or texture). For example, Fig.~\ref{fig:teaser} row 1 show two examples of Local Edit.
Background Edit means changing the background and remain the main object unchanged. For example, Fig.~\ref{fig:teaser} row 2 column 1 and row 4 column 2 show two examples of Background Edit.
Global Edit transfer the overall style of an image. For example, Fig.~\ref{fig:teaser} row 4 column 1 show one example of Global Edit.
Addition and Remove separately add and remove object from an image. For example, Fig.~\ref{fig:teaser} row 3 show two examples of adding object, and row 2 column 2 shows one example of removing object.
We classify editing category to these 4 categories since they can cover most editing instruction types, and requires different operations in editing.

Then, in step 2, we find the main editing object by making further conversations with GPT4-o based on the editing category. For example, editing instruction ``Make the horse into a unicorn" has a main object of ``horse". For editing categories of global edit and addition that do not contain an editing object, we leave the editing object blank.
After that, we use Grounded-SAM~\cite{groundedsam} combined with GPT4-o to obtain the editing mask in step 3. Specifically, we generate image background as mask for Background Edit. For Local Edit and Removing, we use the main object of step 2 as the input of Grounded-SAM to locate the editing mask. For addition editing category, we use the visual ability of GPT4-o and get a possible location of adding the object. For Global editing, the mask is default as the whole image.

Finally, in step 4, we use image generation model to perform image editing based on a target prompt generated by GPT4-o. For Background Edit, Local Edit, Remove, and Addition, we use image inpainting model BrushNet~\cite{brushnet} combined with PowerPaint~\cite{powerpaint} to inpaint the masked region based on the target prompt. For Global Edit, we use InfEdit~\cite{infedit} to change the global style. 
We find that a direct utilization of the diffusion model by giving conditions of mask and text prompt can fully leverage the capabilities of these models, thus leads to high-quality generation results.

\section{Experiment Results}

\begin{table}[htbp]
\centering
\renewcommand\arraystretch{0.8}
\small
\setlength{\tabcolsep}{0.8mm}{
\begin{tabular}{lcccc}
\toprule
\multirow{2}{*}{Method}    & \multirow{2}{*}{Rank}   & Edit  & Content  & Overall Instruction \\ 
  &  & Faithfulness & Preservation &  Following \\ \midrule
IIIE     & 1    & \textbf{0.51}              & \textbf{0.78}                 & \textbf{0.80}                          \\
LEdits++ & 2    & 0.46              & 0.64                 & 0.74                          \\
Tasvir   & 3    & 0.40              & 0.49                 & 0.62                          \\ \bottomrule
\end{tabular}}
\caption{\textbf{Comparison of IIIE and the other two winning solutions.} The score is calculated with an average of 1.2k images in MAGIC benchmark. A bigger score means a better user preference. \textbf{Bold} stands for the best results.}
\label{tab:comparison}
\end{table}

Visualization comparison in Fig.~\ref{fig:teaser} show a higher success rate and visual quality of IIIE compared to previous methods. In the MAGIC workshop, a benchmark of 1.2k images is used for evaluation and 3 winning solutions are announced. The benchmark measure the quality of different editing methods using user study on instruct following, edit fidelity, and content preservation. 3 raters are involved in each job and the majority vote is taken as the final results. Each image is rated with 1 or 0 for each metric. Results of the three winning solution of this competition is shown in Tab.~\ref{tab:comparison}. IIIE show better score on all three metrics compared to the other two winning solutions. We have made our code and edited results publicly available in the hope of our findings can offer some insights for relevant field~\cite{ju2023human,ju2024miradata,liu2022semantic}.

In conclusion, in this technique report presents the winning solution for the CVPR2024 GenAI Media Generation Challenge's Instruction-guided Image Editing track. 
We show that without cumbersome fine-tuning or training, a simple combination of LLMs and text-to-image diffusion model can lead to a image editing results with better performance and higher success rates than previous methods.

{
\small
\bibliographystyle{ieee_fullname}
\bibliography{main}

\begin{thebibliography}{10}\itemsep=-1pt

\bibitem{instructpix2pix}
Tim Brooks, Aleksander Holynski, and Alexei~A Efros.
\newblock Instructpix2pix: Learning to follow image editing instructions.
\newblock In {\em Proceedings of the IEEE/CVF Conference on Computer Vision and Pattern Recognition}, pages 18392--18402, 2023.

\bibitem{masactrl}
Mingdeng Cao, Xintao Wang, Zhongang Qi, Ying Shan, Xiaohu Qie, and Yinqiang Zheng.
\newblock Masactrl: Tuning-free mutual self-attention control for consistent image synthesis and editing.
\newblock In {\em Proceedings of the IEEE/CVF International Conference on Computer Vision}, pages 22560--22570, 2023.

\bibitem{emu}
Xiaoliang Dai, Ji Hou, Chih-Yao Ma, Sam Tsai, Jialiang Wang, Rui Wang, Peizhao Zhang, Simon Vandenhende, Xiaofang Wang, Abhimanyu Dubey, et~al.
\newblock Emu: Enhancing image generation models using photogenic needles in a haystack.
\newblock {\em arXiv preprint arXiv:2309.15807}, 2023.

\bibitem{mgie}
Tsu-Jui Fu, Wenze Hu, Xianzhi Du, William~Yang Wang, Yinfei Yang, and Zhe Gan.
\newblock Guiding instruction-based image editing via multimodal large language models.
\newblock {\em arXiv preprint arXiv:2309.17102}, 2023.

\bibitem{instructdiffusion}
Zigang Geng, Binxin Yang, Tiankai Hang, Chen Li, Shuyang Gu, Ting Zhang, Jianmin Bao, Zheng Zhang, Houqiang Li, Han Hu, et~al.
\newblock Instructdiffusion: A generalist modeling interface for vision tasks.
\newblock In {\em Proceedings of the IEEE/CVF Conference on Computer Vision and Pattern Recognition}, pages 12709--12720, 2024.

\bibitem{prompt2prompt}
Amir Hertz, Ron Mokady, Jay Tenenbaum, Kfir Aberman, Yael Pritch, and Daniel Cohen-Or.
\newblock Prompt-to-prompt image editing with cross attention control.
\newblock {\em arXiv preprint arXiv:2208.01626}, 2022.

\bibitem{ho2020denoising}
Jonathan Ho, Ajay Jain, and Pieter Abbeel.
\newblock Denoising diffusion probabilistic models.
\newblock {\em Advances in Neural Information Processing Systems ({NIPS})}, 33:6840--6851, 2020.

\bibitem{smartedit}
Yuzhou Huang, Liangbin Xie, Xintao Wang, Ziyang Yuan, Xiaodong Cun, Yixiao Ge, Jiantao Zhou, Chao Dong, Rui Huang, Ruimao Zhang, et~al.
\newblock Smartedit: Exploring complex instruction-based image editing with multimodal large language models.
\newblock In {\em Proceedings of the IEEE/CVF Conference on Computer Vision and Pattern Recognition}, pages 8362--8371, 2024.

\bibitem{ju2024miradata}
Xuan Ju, Yiming Gao, Zhaoyang Zhang, Ziyang Yuan, Xintao Wang, Ailing Zeng, Yu Xiong, Qiang Xu, and Ying Shan.
\newblock Miradata: A large-scale video dataset with long durations and structured captions.
\newblock {\em arXiv preprint arXiv:2407.06358}, 2024.

\bibitem{brushnet}
Xuan Ju, Xian Liu, Xintao Wang, Yuxuan Bian, Ying Shan, and Qiang Xu.
\newblock Brushnet: A plug-and-play image inpainting model with decomposed dual-branch diffusion.
\newblock {\em arXiv preprint arXiv:2403.06976}, 2024.

\bibitem{directinversion}
Xuan Ju, Ailing Zeng, Yuxuan Bian, Shaoteng Liu, and Qiang Xu.
\newblock Direct inversion: Boosting diffusion-based editing with 3 lines of code.
\newblock {\em arXiv preprint arXiv:2310.01506}, 2023.

\bibitem{ju2023human}
Xuan Ju, Ailing Zeng, Jianan Wang, Qiang Xu, and Lei Zhang.
\newblock Human-art: A versatile human-centric dataset bridging natural and artificial scenes.
\newblock In {\em Proceedings of the IEEE/CVF Conference on Computer Vision and Pattern Recognition}, pages 618--629, 2023.

\bibitem{ju2023humansd}
Xuan Ju, Ailing Zeng, Chenchen Zhao, Jianan Wang, Lei Zhang, and Qiang Xu.
\newblock Humansd: A native skeleton-guided diffusion model for human image generation.
\newblock In {\em Proceedings of the IEEE/CVF International Conference on Computer Vision}, pages 15988--15998, 2023.

\bibitem{li2024textcraftor}
Yanyu Li, Xian Liu, Anil Kag, Ju Hu, Yerlan Idelbayev, Dhritiman Sagar, Yanzhi Wang, Sergey Tulyakov, and Jian Ren.
\newblock Textcraftor: Your text encoder can be image quality controller.
\newblock In {\em Proceedings of the IEEE/CVF Conference on Computer Vision and Pattern Recognition}, pages 7985--7995, 2024.

\bibitem{liu2023hyperhuman}
Xian Liu, Jian Ren, Aliaksandr Siarohin, Ivan Skorokhodov, Yanyu Li, Dahua Lin, Xihui Liu, Ziwei Liu, and Sergey Tulyakov.
\newblock Hyperhuman: Hyper-realistic human generation with latent structural diffusion.
\newblock {\em arXiv preprint arXiv:2310.08579}, 2023.

\bibitem{liu2022semantic}
Xian Liu, Yinghao Xu, Qianyi Wu, Hang Zhou, Wayne Wu, and Bolei Zhou.
\newblock Semantic-aware implicit neural audio-driven video portrait generation.
\newblock In {\em European conference on computer vision}, pages 106--125. Springer, 2022.

\bibitem{groundedsam}
Tianhe Ren, Shilong Liu, Ailing Zeng, Jing Lin, Kunchang Li, He Cao, Jiayu Chen, Xinyu Huang, Yukang Chen, Feng Yan, et~al.
\newblock Grounded sam: Assembling open-world models for diverse visual tasks.
\newblock {\em arXiv preprint arXiv:2401.14159}, 2024.

\bibitem{Rombach_2022_CVPR}
Robin Rombach, Andreas Blattmann, Dominik Lorenz, Patrick Esser, and Bj\"orn Ommer.
\newblock High-resolution image synthesis with latent diffusion models.
\newblock In {\em Proceedings of the IEEE/CVF Conference on Computer Vision and Pattern Recognition ({CVPR})}, pages 10684--10695, June 2022.

\bibitem{song2020denoising}
Jiaming Song, Chenlin Meng, and Stefano Ermon.
\newblock Denoising diffusion implicit models.
\newblock {\em arXiv preprint arXiv:2010.02502}, 2020.

\bibitem{infedit}
Sihan Xu, Yidong Huang, Jiayi Pan, Ziqiao Ma, and Joyce Chai.
\newblock Inversion-free image editing with natural language.
\newblock {\em arXiv preprint arXiv:2312.04965}, 2023.

\bibitem{powerpaint}
Junhao Zhuang, Yanhong Zeng, Wenran Liu, Chun Yuan, and Kai Chen.
\newblock A task is worth one word: Learning with task prompts for high-quality versatile image inpainting.
\newblock {\em arXiv preprint arXiv:2312.03594}, 2023.

\end{thebibliography}
}

\end{document}